\documentclass[a4paper]{article}

\usepackage{graphicx}

\usepackage{amsmath,amssymb}

\usepackage{cite}

\usepackage{url}

\usepackage{color}

\usepackage{multirow}

\usepackage{booktabs}

\usepackage{titlesec}

\usepackage{enumitem}

\usepackage[T1]{fontenc}

\usepackage{lmodern}

\usepackage[none]{hyphenat}

\usepackage[left=2cm,right=2cm,top=2cm,bottom=2cm]{geometry}

\usepackage{ragged2e}

\justifying

\graphicspath{{./}}

\title{\bfseries\LARGE Toward Dignity-Aware AI: Next-Generation Elderly Monitoring from Fall Detection to ADL}

\author{Xun Shao$^{1}$, Aoba Otani$^{1}$, Yuto Hirasuka$^{1}$, Runji Cai$^{2}$, Seng W. Loke$^{3}$\\ \small $^{1}$Toyohashi University of Technology\; $^{2}$Guangdong Ocean University\; $^{3}$Deakin University}

\date{}

\begin{document}

\maketitle

\begin{abstract}

This position paper envisions a next-generation elderly monitoring system that moves beyond fall detection toward the broader goal of Activities of Daily Living (ADL) recognition. Our ultimate aim is to design privacy-preserving, edge-deployed, and federated AI systems that can robustly detect and understand daily routines, supporting independence and dignity in aging societies. At present, ADL-specific datasets are still under collection. As a preliminary step, we demonstrate feasibility through experiments using the SISFall dataset and its GAN-augmented variants, treating fall detection as a proxy task. We report initial results on federated learning with non-IID conditions, and embedded deployment on Jetson Orin Nano devices. We then outline open challenges such as domain shift, data scarcity, and privacy risks, and propose directions toward full ADL monitoring in smart-room environments. This work highlights the transition from single-task detection to comprehensive daily activity recognition, providing both early evidence and a roadmap for sustainable and human-centered elderly care AI.

\end{abstract}

\section{Introduction and Related Work}

\subsection{Motivation}

The rapid demographic shift toward aging societies has intensified the urgency of developing intelligent, privacy-preserving monitoring systems to support independent living for older adults. According to the World Health Organization (WHO), falls are one of the leading causes of injury-related mortality among elderly populations, with nearly $37.3$ million falls requiring medical intervention each year \cite{WHO2021Falls}. However, falls are only one dimension of Activities of Daily Living (ADL), which include essential routines such as mobility, sleep, toileting, eating, and personal hygiene \cite{Katz1963ADL}. Continuous monitoring of ADL not only enables timely detection of critical events such as falls, but also provides early indicators of frailty, cognitive decline, or deteriorating health status \cite{ravi2017deep, Wang2021ADLBinaryEnvSensors}. This paper advocates for a shift in focus: from isolated fall detection to holistic ADL recognition, supported by edge AI and Federated Learning (FL).

\subsection{Fall Detection as a Proxy Task}

Despite the importance of ADL monitoring, most existing studies have focused on fall detection. This is partly because falls are acute and life-threatening events, and partly because benchmark datasets exist. For instance, the SISFall dataset provides annotated accelerometer and gyroscope data for simulated falls \cite{Guerrero2019SISFall}. In this work, we emphasize that fall detection is not our ultimate goal. Instead, we use it as a proxy task to validate algorithms and system components in the absence of large-scale ADL datasets. All preliminary experiments in this paper are conducted solely on SISFall and its GAN-augmented variants. ADL-specific datasets are still under collection in our smart-room testbed.

\subsection{Approaches to Fall Detection}

\textbf{Wearables:} Accelerometer and gyroscope sensors embedded in smartphones or body-worn devices are among the most widely studied for fall detection \cite{mubashir2013fallsurvey, Chen2020WearableHAR}. These approaches are generally accurate in controlled conditions. However, they face practical challenges: elderly users may resist wearing sensors, forget to recharge devices, or experience discomfort. Moreover, continuous usage can be intrusive in daily life.\\

\textbf{Vision-based:} Camera systems, including depth sensors (e.g., Microsoft Kinect), have been applied for fall recognition by analyzing human poses and motion trajectories \cite{Mastorakis2014VisionBased, Stone2015KinectFalls}. Although vision-based systems can achieve high accuracy, they raise serious privacy concerns in sensitive environments such as nursing homes or private homes. In addition, performance may degrade under poor lighting or occlusion, and deployment often requires expensive infrastructure.\\

\textbf{Device-free RF:} Recent advances in device-free sensing exploit wireless signals, such as WiFi Channel State Information (CSI), Ultra-wideband (UWB), or mmWave radar, to detect falls without requiring wearables or cameras \cite{Wang2017DeviceFree, Adib2015RFTrack, yang2020uwb, zhao2018mfd}. These approaches offer privacy advantages and can penetrate occlusions. However, they often require specialized hardware and are highly sensitive to environmental variations, limiting their scalability.

\subsection{Toward ADL Recognition}

While fall detection is a critical starting point, it does not capture the full picture of elderly well-being. ADL monitoring enables a more comprehensive understanding of daily routines, such as whether an elderly person gets out of bed, visits the bathroom, or remains inactive for extended periods \cite{Camp2021ADLTechScopingReview, ignatov2018cnn}. Deviations in such patterns may indicate emerging health risks. Thus, our vision is to extend beyond fall detection to a holistic ADL monitoring framework. We stress that our position paper proposes ADL recognition as the ultimate goal, with fall detection serving as an initial case study.

\subsection{Challenges: Data Scarcity and Generalization}

A central obstacle is the scarcity of real-world ADL datasets. Data collection in elderly populations raises ethical, privacy, and logistical challenges. While fall datasets such as SISFall exist, large-scale ADL datasets remain rare. To mitigate this, Generative Adversarial Networks (GANs) have been adopted for data augmentation, synthesizing rare falls or subtle transitions to improve class balance and generalization \cite{Li2021MetaHAR, Yoon2019TimeGAN}. Nevertheless, synthetic data cannot fully replace real-world ADL data, underscoring the need for continued dataset development.

Generalization across heterogeneous environments also poses challenges. Models trained in one home or facility may not transfer well to others due to differences in sensor placement, architecture, or individual behavior. Addressing domain shift requires techniques such as Continual Learning and domain adaptation \cite{delange2021continual, zhuang2020survey}.

\subsection{Edge AI and Federated Learning}

Privacy concerns further constrain centralized data collection. Federated Learning (FL) provides a paradigm where each device trains locally and shares only model parameters, preserving privacy while enabling collective learning \cite{mcmahan2017communication, Li2020FLChallenges}. Applications of FL in healthcare demonstrate its potential to balance personalization and global generalization \cite{Li2021MetaHAR, Chen2020FedHealth, Zhou2022TwoDFLPersonalizedHAR}. Moreover, deploying models on embedded devices such as NVIDIA Jetson Orin Nano enables low-latency, energy-efficient inference at the edge \cite{howard2017mobilenets, han2016deeplearningcompression}. In our preliminary experiments, quantized Conv-GRU models achieved efficient inference on Jetson hardware.

\subsection{Our Position and Contributions}

In summary, our position paper makes the following contributions:

\begin{itemize}[leftmargin=*,noitemsep]

  \item We present a vision of an ADL-centered, privacy-preserving monitoring system integrating multimodal sensing, Edge AI, and Federated Learning.

  \item We clarify that all current results are based solely on SISFall and its GAN-augmented extensions, using fall detection as a proxy task.

  \item We review existing approaches to fall detection and argue for the necessity of moving beyond falls toward ADL recognition.

  \item We identify open challenges in data scarcity, domain shift, privacy, and sensor heterogeneity, and propose directions toward holistic ADL monitoring in smart-room environments.

\end{itemize}

\section{Vision and Conceptual Framework}

\begin{figure}[htbp]

  \centering

  \includegraphics[width=0.95\linewidth]{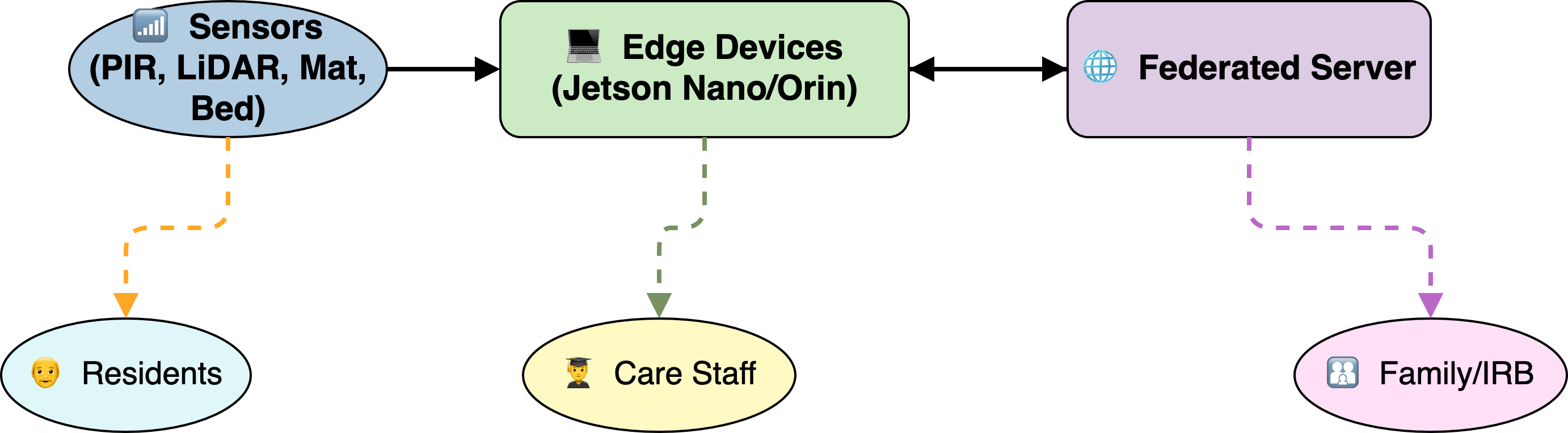}

  \caption{Conceptual system-level overview of the proposed federated monitoring framework. Current experiments use SISFall IMU data only. Other modalities (e.g., LiDAR, pressure mats) are part of our planned smart-room deployments.}

  \label{fig:system}

\end{figure}

Our ultimate aim is ADL-centric elderly monitoring that preserves privacy while enabling robust, on-device intelligence. Fall detection is used only as a proxy task in our preliminary experiments; the broader vision is to monitor a wide range of ADL such as mobility, sleeping, and bathroom visits. Figure \ref{fig:system} illustrates the conceptual system design.

\subsection{System Overview}

The envisioned system integrates three pillars:

\begin{itemize}[leftmargin=*,noitemsep]

  \item Multi-modal sensing in a smart-room (IMU, pressure mats, and LiDAR).

  \item Edge AI inference on embedded devices such as NVIDIA Jetson Orin Nano, for low-latency and privacy-preserving local processing.

  \item Federated Learning (FL) to collaboratively train global models without centralizing raw data \cite{mcmahan2017communication, Li2020FLChallenges}.

\end{itemize}

While only IMU-derived SISFall data are used in current results, the architecture is designed to accommodate additional modalities once real ADL datasets are available.

\subsection{Smart-Room Testbed (Built; Data Collection Ongoing)}

\begin{figure}[htbp]

  \centering

  \includegraphics[width=0.9\linewidth]{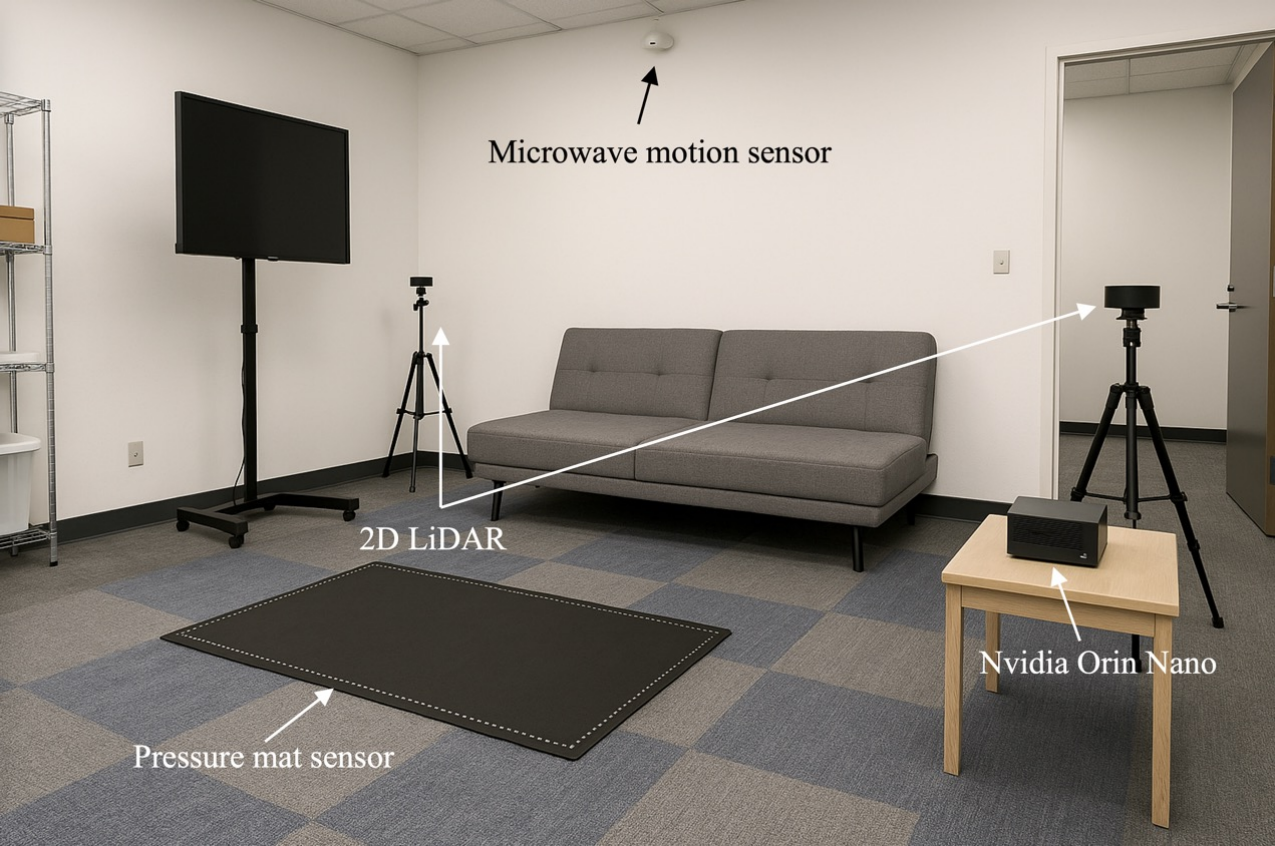}

  \caption{Conceptual elderly monitoring testbed with anonymous sensors at Toyohashi University of Technology (TUT). Data from non-identifiable modalities such as floor pressure, LiDAR, and motion sensors are envisioned to be processed on local Edge AI devices deployed at the room level. Local anomaly detection and activity models can be trained on-device, while federated learning aggregates only model parameters $(\theta)$ in the cloud without transmitting raw personal data, thereby enabling privacy-preserving global improvement.}

  \label{fig:room}

\end{figure}

Our smart-room testbed at Toyohashi University of Technology has been fully built and instrumented as shown in Fig. \ref{fig:room}. The environment integrates privacy-friendly, non-visual sensors such as pressure mats, microwave motion sensors, and LiDAR as future extensions. As of this paper, ADL data collection in the smart room is ongoing, and no smart-room data are included in the results reported here.

\subsection{Modeling Pipeline (Conceptual)}

\begin{figure}[htbp]

  \centering

  \includegraphics[width=0.92\linewidth]{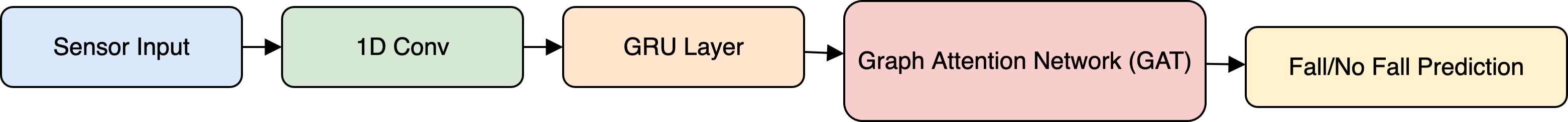}

  \caption{Conceptual spatiotemporal pipeline with Conv-GRU and planned GAT fusion. In this position paper, GAT is a planned extension for multi-sensor ADL datasets; current experiments use Conv-GRU on SISFall IMU data only.}

  \label{fig:convgrugat}

\end{figure}

Sensor signals are processed by a convolutional GRU (Conv-GRU) module to capture temporal dynamics. A Graph Attention Network (GAT) is conceptually layered on top to learn inter-sensor dependencies (Fig. \ref{fig:convgrugat}). In current experiments, only SISFall IMU signals are used; multi-sensor GAT fusion will be evaluated once ADL datasets are collected.

\subsection{Synthetic Data and Data Scarcity}

\begin{figure}[htbp]

  \centering

  \includegraphics[width=0.85\linewidth]{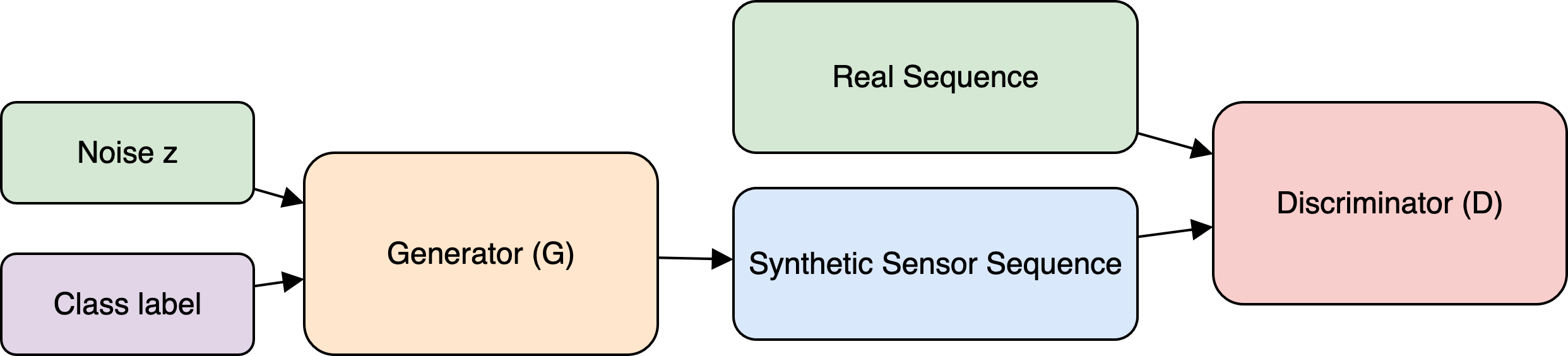}

  \caption{cGAN architecture for generating synthetic SISFall sequences (IMU only).}

  \label{fig:cgan}

\end{figure}

To address class imbalance, we employ conditional GANs (cGANs) trained on SISFall to synthesize rare fall patterns (Figure~\ref{fig:cgan}). While the figure depicts the general GAN setup, note that the current experiments focus exclusively on IMU-based SISFall inputs, and multi-modal GAN extensions will be explored in future smart-room studies.

\subsection{Embedded Deployment (Planned and Partial)}

All models are designed for embedded deployment on Jetson Orin Nano devices. Our preliminary benchmarks (Section~\ref{sec:jetson}) confirm feasibility for SISFall-based Conv-GRU. INT$8$ quantization with TensorRT is used for efficient inference \cite{howard2017mobilenets, han2016deeplearningcompression}. Actual latency and model size measurements are reported in Section~\ref{sec:jetson}.

\section{Preliminary Results}

All results in this section are based solely on the SISFall dataset and its GAN-augmented variants. The smart-room testbed is built and currently collecting ADL data, but no smart-room data are used in this paper. We report (i) GAN-based augmentation on SISFall IMU streams, (ii) robustness under IID/Non-IID simulations, and (iii) embedded inference on Jetson Orin Nano.

\subsection{Common Setup}

Unless otherwise noted, Conv-GRU is used with window length $T=100$, Adam $(lr= 0.001)$, batch size $32$, and $100$ FL rounds with FedAvg (local epochs = $3$) across $K=5$ clients. Metrics are macro-F$1$, accuracy, and FPR on a held-out test split. All curves report mean$\pm$std over three random seeds.

\subsection{SISFall and GAN-Synthesized Extensions}

\begin{figure}[htbp]

  \centering

  \includegraphics[width=0.7\linewidth]{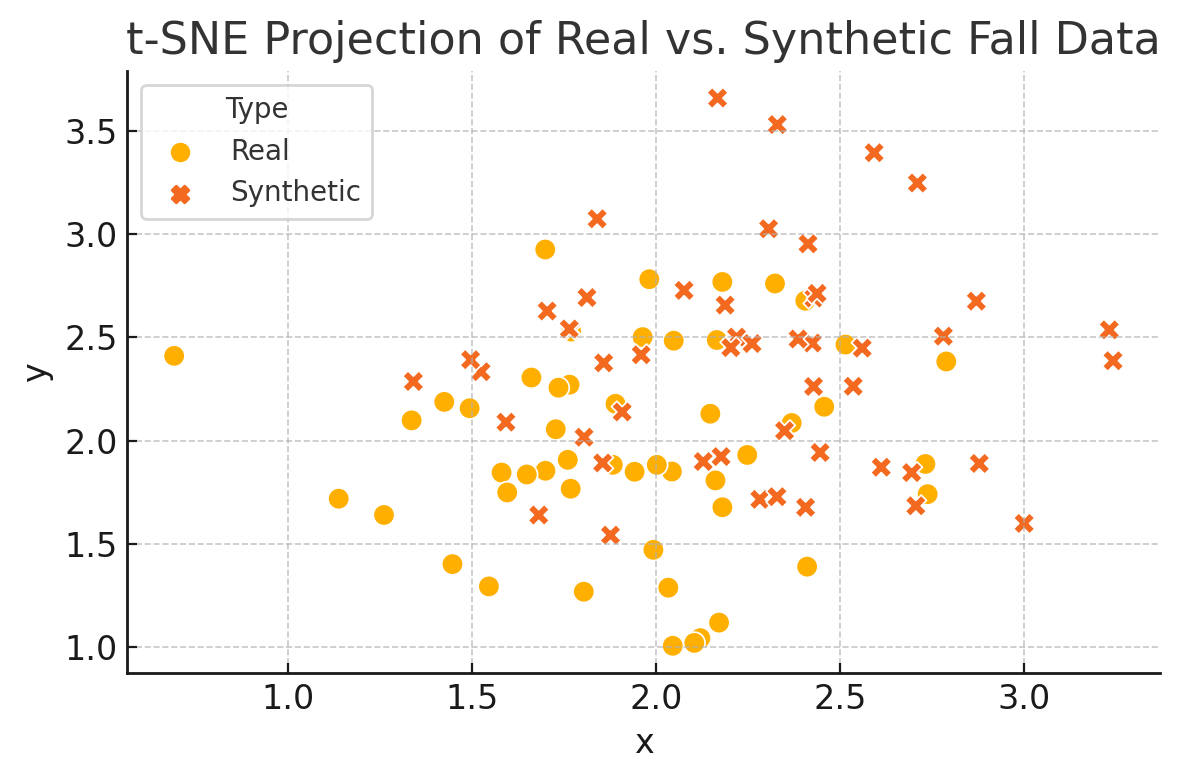}

  \caption{t-SNE visualization comparing real and GAN-synthesized SISFall data distributions. The close overlap indicates that synthetic sequences achieve high fidelity relative to real data.}

  \label{fig:tsne}

\end{figure}

To address data imbalance, we trained conditional GANs (cGANs) on SISFall fall sequences. This enabled synthesis of rare fall types and subtle transitions, enriching the dataset. Figure \ref{fig:tsne} shows a t-SNE visualization of real versus synthetic distributions, demonstrating good alignment in feature space. Note: LiDAR-derived synthetic inputs are excluded; only IMU-based SISFall data were used.

\begin{figure}[htbp]

  \centering

  \includegraphics[width=0.7\linewidth]{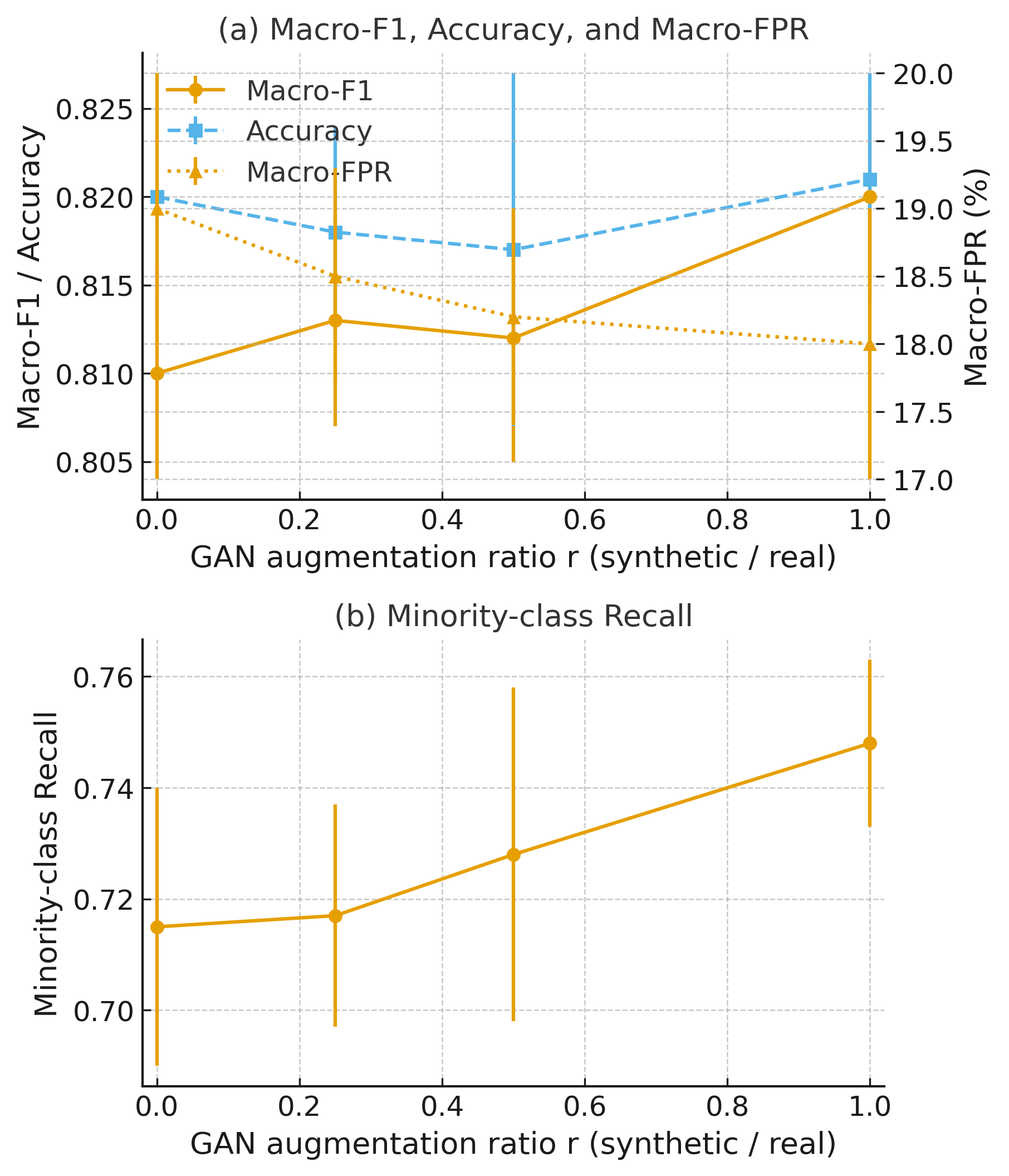}

  \caption{Effect of GAN augmentation ratio $r$ (synthetic/real) on SISFall performance (mean$\pm$std over $3$ seeds). (a) Macro-F$1$, Accuracy, and Macro-FPR. (b) Minority-class recall steadily improves with higher $r$, suggesting the potential benefit of GAN-based augmentation for rare-class detection.}

  \label{fig:GAN_ratio}

\end{figure}

We varied the GAN augmentation ratio $r \in 0, 0.25, 0.5, 1.0$, defined as the number of synthetic samples relative to real training samples per class. Fig. \ref{fig:GAN_ratio} summarizes the results. Panel (a) shows overall performance metrics (Macro-F$1$, Accuracy, and Macro-FPR), which remain stable across augmentation levels. Panel (b) highlights minority-class recall, which improves steadily with higher augmentation ratios, indicating that GAN-synthesized samples help balance rare fall events without degrading overall accuracy.

\subsection{Model Evaluation Under IID and Non-IID}

We consider several candidate architectures for future ADL-oriented deployments, including Conv-GRU (baseline), GRU-only, CNN-LSTM, and a planned Conv-GRU+GAT extension for multi-sensor fusion. In this position paper, however, all quantitative experiments are conducted on the SISFall IMU proxy task using the Conv-GRU baseline, with GAN-augmented samples included. Evaluation considered two federated conditions: (i) IID (random partitions) and (ii) Non-IID (cross-partition to simulate inter-site heterogeneity). Table~\ref{tab:table1} summarizes representative results.

\begin{table}[htbp]

  \centering

  \caption{Preliminary SISFall Conv-GRU performance with GAN-augmented data under IID and Non-IID federated partitions.}

  \label{tab:table1}

  \begin{tabular}{lcccc}

    \toprule

    Condition & Accuracy & F$1$ Score & FPR & Comm. Overhead \\\\

    \midrule

    IID & $95.2$\% & $0.94$ & $3.1$\% & Low \\\\

    Non-IID & $88.7$\% & $0.89$ & $7.5$\% & Medium \\\\

    \bottomrule

  \end{tabular}

\end{table}

IID implies random partitioning within SISFall; while Non-IID is subject-wise label-skew simulated via a Dirichlet prior with $\alpha = 0.5$ across $K=5$ clients (smaller $\alpha$ implies stronger skew).

\begin{figure}[htbp]

  \centering

  \includegraphics[width=1.0\linewidth]{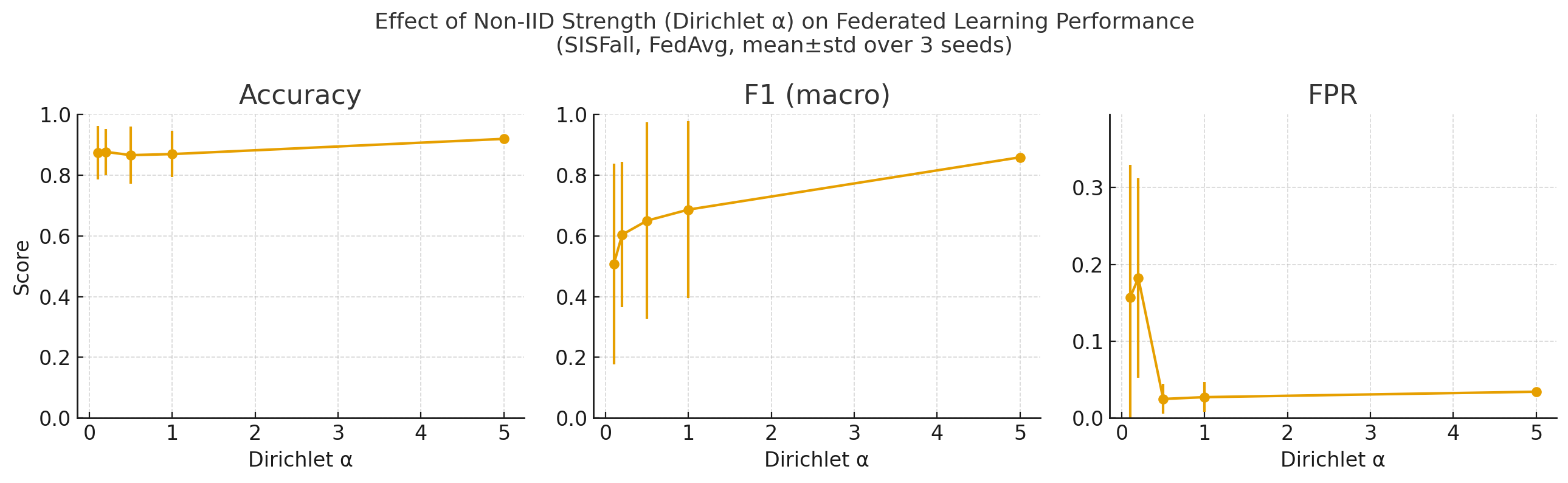}

  \caption{Impact of non-IID severity (Dirichlet $\alpha$) on SISFall performance with GAN augmentation (mean$\pm$std over $3$ seeds). Results are shown for (left) Accuracy, (center) Macro-F$1$, and (right) Macro-FPR. Smaller $\alpha$ indicates stronger label skew across clients, which leads to reduced Accuracy and Macro-F$1$ and increased Macro-FPR, confirming the negative impact of data heterogeneity in federated learning.}

  \label{fig:dirichlet}

\end{figure}

Label skew across clients is controlled by a Dirichlet prior with $\alpha \in \{0.1,0.2,0.5,1.0,5.0\}$ for $K=5$ clients (smaller $\alpha$ increases skew). For each $\alpha$, we run FedAvg for $100$ rounds (local epochs= $3$, batch size= $32$, lr = $0.001$) and report the global model’s macro-F$1$, accuracy, and FPR as mean$\pm$std over three random seeds (Fig. \ref{fig:dirichlet}).

\subsection{Evaluation Framework}

\begin{figure}[htbp]

  \centering

  \includegraphics[width=0.6\linewidth]{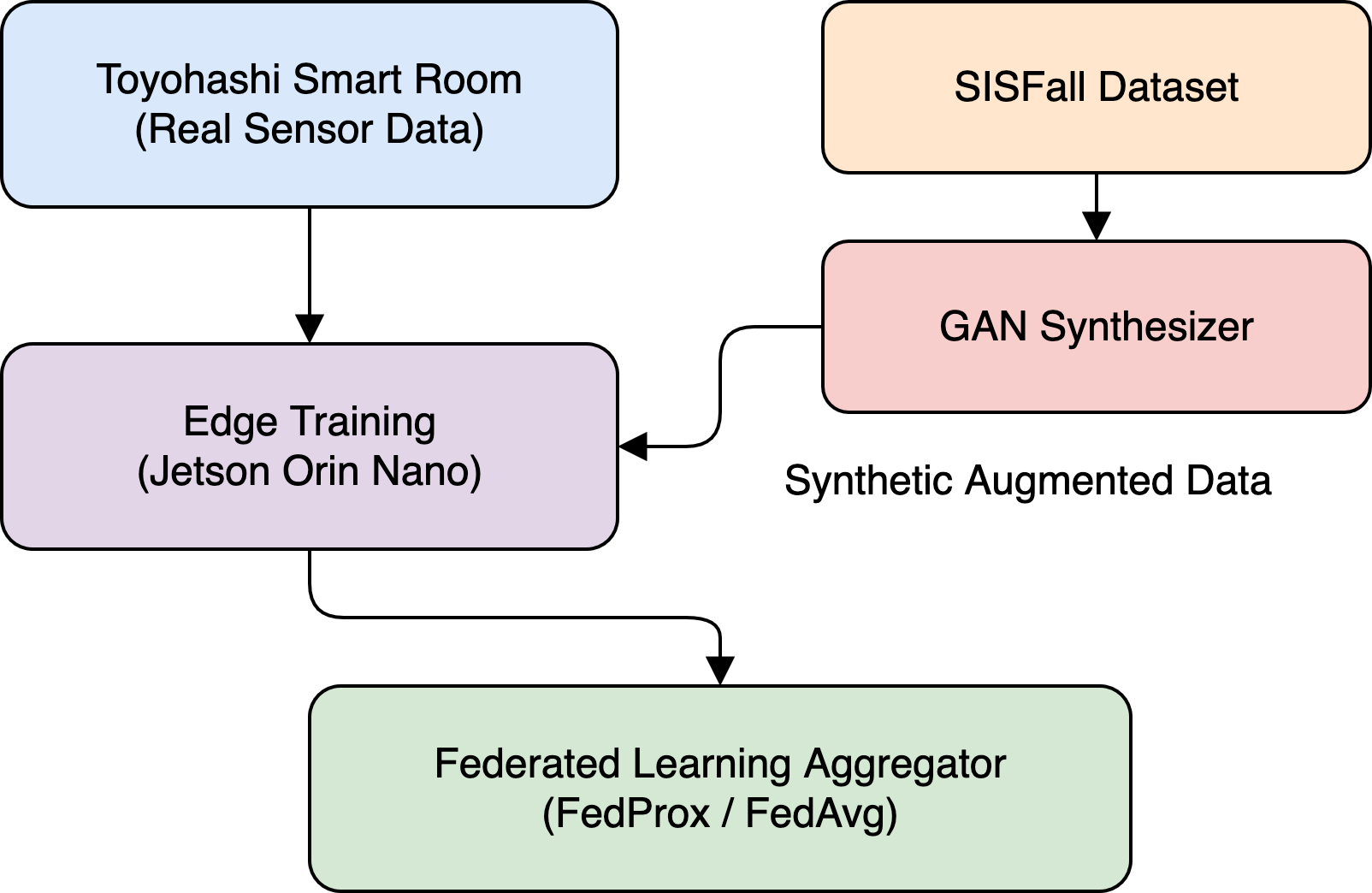}

  \caption{Evaluation framework combining the built smart-room testbed (ADL data collection ongoing; not used in the current results), SISFall (with GAN-based augmentation), and the edge/FL pipeline. This diagram bridges the proxy-task results in this paper and future ADL-oriented evaluations, while explicitly separating current experiments (SISFall only) from planned smart-room ADL studies.}

  \label{fig:eval_framework}

\end{figure}

To clarify how proxy-task experiments relate to future ADL deployments, we restore the evaluation framework diagram. It summarizes the data flow between the smart-room testbed (data collection ongoing), the SISFall dataset with GAN-based synthesis, and the edge/FL pipeline used for our experiments (Fig. \ref{fig:eval_framework}).

\subsection{Embedded Deployment on Jetson Orin Nano}

\label{sec:jetson}

\begin{figure}[htbp]

  \centering

  \includegraphics[width=0.7\linewidth]{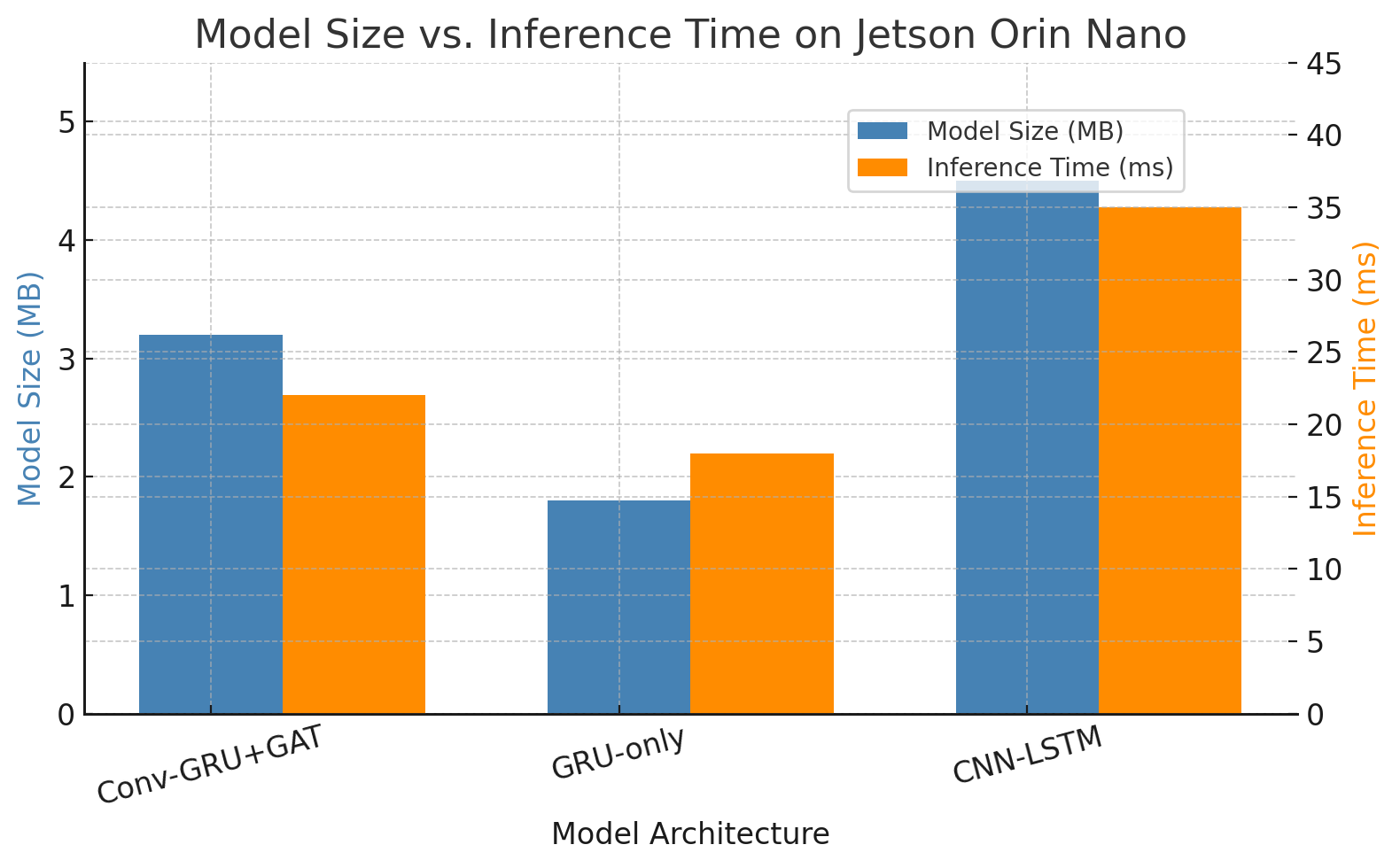}

  \caption{Latency and model size comparison on Jetson Orin Nano (INT$8$ quantization). Results are measured for the SISFall Conv-GRU baseline. Additional bars (e.g., CNN-LSTM and Conv-GRU+GAT) are shown as projected complexity references for future multimodal extensions, and are not part of the current experimental evaluation.}

  \label{fig:nano}

\end{figure}

We deployed the Conv-GRU baseline (without multimodal extensions) on a Jetson Orin Nano device. Models were quantized to INT$8$ using TensorRT. Fig. \ref{fig:nano} shows that real-time inference is achievable for the Conv-GRU baseline within the device’s low-power operating mode. For completeness, projected latency/model-size references for planned extensions (e.g., Conv-GRU+GAT) are also included in the figure, although they are not evaluated in the current SISFall experiments.

\subsection{Scope and Limitations of Results}

These results demonstrate feasibility of (i) GAN-based data augmentation, (ii) federated-style evaluation under Non-IID conditions, and (iii) embedded inference. However, it is crucial to note that:

\begin{itemize}[leftmargin=*,noitemsep]

  \item Data: Only SISFall IMU data (plus GAN augmentation) were used.

  \item Sensors: LiDAR, pressure mats, and motion sensors have not yet been tested.

  \item Testbed: The smart room is built and ADL data collection is ongoing; no smart-room data are included in these results.

\end{itemize}

Therefore, the above results should be regarded strictly as preliminary, with future validation required on real ADL datasets.

\section{Discussion and Future Challenges}

Our preliminary results based on SISFall and its GAN-augmented variants confirm that edge-deployed sequence models (e.g., Conv-GRU) can achieve real-time feasibility on embedded devices and maintain robustness under simulated Non-IID conditions, while robustness to sensor dropout remains future work. These findings are promising, but they also highlight the limitations of evaluating on a single proxy dataset. The next stage of our research is to bridge from fall detection on SISFall to holistic ADL recognition in real smart-room deployments. In this section, we outline the key challenges and future directions, linking them directly to our preliminary results.

\subsection{From Proxy Task to ADL Monitoring}

While SISFall allows us to test algorithms and deployment feasibility, it does not cover the full spectrum of ADL behaviors (e.g., bed exit/entry, prolonged inactivity, toileting patterns). Thus, our GAN-augmented results should be viewed as methodological validation rather than final benchmarks. Future experiments will require multimodal ADL datasets currently under collection in our smart-room testbed. In this regard, the methodological insights suggested by our figures provide useful guidance for future ADL deployments:

\begin{itemize}[leftmargin=*,noitemsep]

  \item Figure \ref{fig:GAN_ratio} (GAN augmentation): The observed gains in minority-class recall highlight the potential of generative approaches to handle rare and subtle ADL patterns, such as irregular toileting or fragmented sleep, which are difficult to capture in small datasets.

  \item Figure \ref{fig:dirichlet} (Non-IID impact): The degradation under strong label skew mirrors the heterogeneity across different elderly living environments. This finding motivates the integration of personalized FL and continual learning for cross-site ADL generalization.

\end{itemize}

By linking these proxy-task results to ADL-oriented challenges, our work moves beyond technical validation and aligns with the broader agenda of supporting independence and dignity in aging societies. More broadly, this framing reinforces that fall detection should be understood as an intermediate step toward holistic Agetech solutions, rather than an end goal in itself.

\subsection{Label Scarcity and Weak Supervision}

GAN-based augmentation improved class balance in SISFall, but real ADL data will face even greater label scarcity, as events such as falls or abnormal ADL deviations are rare and difficult to annotate. Future work will explore self-supervised and semi-supervised learning, reducing reliance on expert-labeled data.

\subsection{Cross-Site Generalization and Domain Shift}

Our Non-IID experiments on SISFall partitions highlight the difficulty of cross-site generalization. In practice, differences in room layouts, sensor placements, and individual behaviors will induce significant domain shift. Continual learning and transfer learning strategies will be critical to adapt models across sites and over time \cite{delange2021continual, zhuang2020survey}.

\subsection{Federated Optimization Trade-offs}

Our results confirm that FL-style training is feasible, but Non-IID data conditions complicate convergence. FedProx-style stabilization is one option, but it introduces additional computation at the edge. Future work must consider communication-efficient aggregation (e.g., sparsification, partial model updates) and personalized FL strategies that balance global accuracy with individual adaptation.

\subsection{Privacy and Ethical Considerations}

Although our preliminary pipeline ensures that raw SISFall data remain local, gradients themselves may leak information through inference attacks. This issue will become more pressing once real ADL data from smart rooms are used. Differential privacy, secure aggregation, and encrypted updates must be integrated into future FL deployments to safeguard user dignity and comply with healthcare regulations.

\begin{figure}[htbp]

  \centering

  \includegraphics[width=0.7\linewidth]{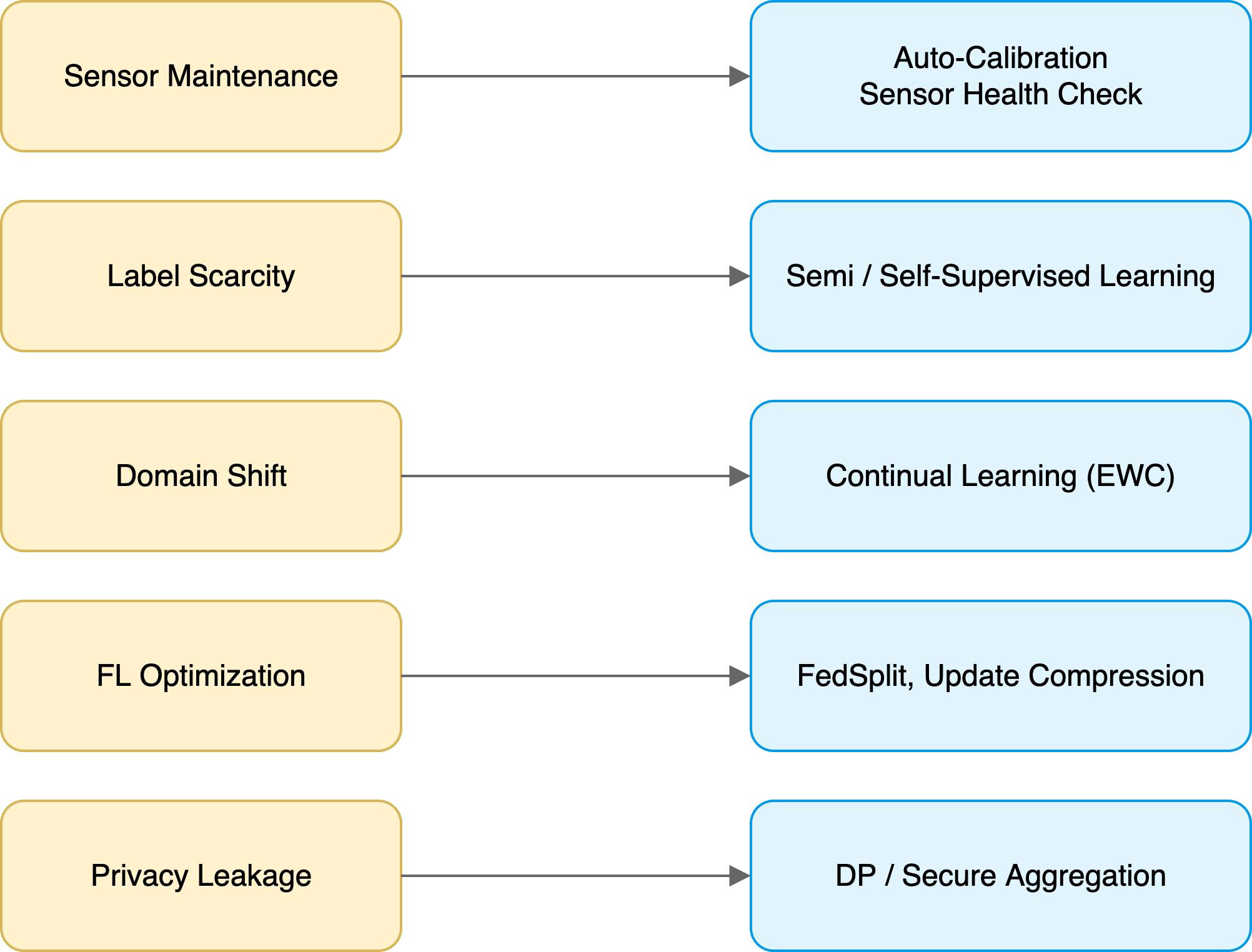}

  \caption{Overview of open challenges for transitioning from fall detection to ADL recognition. Each challenge is linked to preliminary results (SISFall+GAN) and extended to future deployment scenarios.}

  \label{fig:challenges}

\end{figure}

We summarize the challenges with Fig. \ref{fig:challenges}

\subsection{Summary of Research Roadmap}

In summary, our roadmap progresses from: (1) SISFall+GAN proxy validation (this paper); (2) smart-room ADL data collection (ongoing); (3) methodological extensions (continual/domain adaptation/privacy-preserving FL); (4) real-world longitudinal evaluation.

This staged approach ensures that the insights from our preliminary results directly inform the path toward full ADL monitoring systems.

\section{Conclusion}

In this position paper, we have argued that elderly monitoring research must evolve from isolated fall detection toward holistic ADL recognition. While fall detection remains a critical application, our long-term vision is to monitor daily routines such as sleeping, mobility, and toileting, providing early indicators of frailty and supporting independent living with dignity.

Our preliminary results, based solely on the SISFall dataset and its GAN-augmented variants, demonstrate three key aspects: (i) the feasibility of GAN-based data augmentation for balancing rare fall events, (ii) robustness of sequence models under Non-IID conditions, and (iii) real-time inference capability on embedded Jetson Orin Nano devices after quantization. These results validate our methodological approach but must be seen as a proxy-task demonstration rather than final outcomes. No ADL-specific datasets have yet been used, and multimodal sensing (e.g., LiDAR, motion sensors, pressure mats) remains future work.

We have also outlined a staged roadmap:

\begin{itemize}[leftmargin=*,noitemsep]

  \item Stage 1: Proxy-task validation on SISFall (completed).

  \item Stage 2: Ongoing collection of multimodal ADL datasets in smart-room testbeds.

  \item Stage 3: Methodological extensions including continual learning, domain adaptation, and privacy-preserving FL.

  \item Stage 4: Real-world longitudinal evaluation in elderly care facilities.

\end{itemize}

By making explicit the separation between what has been demonstrated and what is planned, this paper provides clarity on both the current evidence and the challenges that lie ahead. We argue that the convergence of Edge AI, Federated Learning, and semantic communication-inspired data abstraction will enable sustainable, privacy-conscious, and human-centered ADL monitoring systems for aging societies.

Ultimately, this position paper serves as both an early evidence base and a research agenda, guiding the transition from fall detection to comprehensive ADL recognition.

\section*{Acknowledgment}

This work acknowledges institutional and funding support (Toyohashi City grant Reiwa 7 (2025), JSPS KAKENHI 24K14913, Naito Foundation 2024--2025, Telecommunications Advancement Foundation (SCAT), and RIEC Tohoku University). We thank the Toyohashi Municipal General Elderly Home (Tsutsuji-so) and volunteers.

\bibliographystyle{unsrt}

\bibliography{references}

\end{document}